\documentclass{article}

\usepackage{PRIMEarxiv}

\usepackage[utf8]{inputenc} 
\usepackage[T1]{fontenc}    
\usepackage{hyperref}       
\usepackage{url}            
\usepackage{booktabs}       
\usepackage{amsfonts}       
\usepackage{nicefrac}       
\usepackage{microtype}      
\usepackage{lipsum}
\usepackage{fancyhdr}       
\usepackage{graphicx}       
\graphicspath{{media/}}     
\usepackage{multirow}
\usepackage{multicol}
\usepackage[numbers]{natbib}
\title{Periodic-MAE: Periodic Video Masked Autoencoder for rPPG Estimation}

\author{
  Jiho Choi and Sang Jun Lee\thanks{Corresponding author.}  \\
Division of Electronics and Information Engineering, \\
Jeonbuk National University, Republic of Korea \\
  \texttt{\{jihochoi, sj.lee\}@jbnu.ac.kr} \\
}

\begin{document}
\maketitle

\begin{abstract}
In this paper, we propose Periodic-MAE, a self-supervised framework for learning generalizable spatio-temporal representations of periodic physiological signals from unlabeled facial videos.
The proposed method leverages a masked autoencoder (MAE), which learns high-dimensional facial representations by reconstructing masked video tokens without relying on remote photoplethysmography (rPPG) specific supervision.
To explicitly align representation learning with the characteristics of rPPG, we introduce a periodicity-aware frame masking strategy based on video resampling, enabling the encoder to learn representations that capture quasi-periodic temporal patterns relevant to pulse signal estimation.
In addition, physiological bandlimit constraints are integrated into the MAE pre-training framework, exploiting the sparsity of pulse signals in the frequency domain to guide the learned representations toward physiologically meaningful patterns.
After pre-training, the learned representations are transferred to downstream rPPG estimation, where the encoder serves as a generic feature extractor for recovering pulse-related signals from facial videos.
We conduct extensive experiments on four benchmark datasets, including PURE, UBFC-rPPG, MMPD, and V4V.
Moreover, we evaluate the proposed approach on a real-world rPPG dataset collected under unconstrained lighting conditions and subject motion.
Experimental results demonstrate that Periodic-MAE consistently improves rPPG estimation performance, particularly in challenging cross-dataset and real-world evaluation settings.
Our code is available at \href{https://github.com/ziiho08/Periodic-MAE}{https://github.com/ziiho08/Periodic-MAE}.
\end{abstract}

\section{Introduction}
\label{introduction}
Remote photoplethysmography (rPPG) is a non-contact method that utilizes a camera sensor to measure physiological signals.
Monitoring physiological signals such as heart rate (HR) and respiration rate is essential for assessing human vital states.
These signals are derived from the blood volume pulse.
Photoplethysmography (PPG) is a non-invasive optical method for detecting blood volume changes. 
However, the PPG sensor requires direct contact with the skin, which can be uncomfortable for infants and patients with skin disorders, and their use is often limited outside clinical settings.
The PPG mechanism relies on measuring variations in light scattering and absorption within subcutaneous blood vessels as the heart contracts and relaxes.
These pulsatile blood volume fluctuations induce subtle changes in the facial skin region that are imperceptible to the human eye but can be captured by a camera, thereby providing the physical foundation for rPPG.
Unlike traditional contact sensors, rPPG does not require physical contact and can be applied in various fields such as remote healthcare, affective recognition, driver monitoring systems, and anti-spoofing.

Early studies of rPPG estimation primarily relied on traditional signal processing techniques, such as color change analysis in facial videos~\citep{balakrishnan2013detecting, poh2010non} and color subspace conversion approaches~\citep{de2013robust, wang2016algorithmic}.
However, the human face is not an ideal Lambertian diffuse surface, and strong noise from illumination variations and subject motion poses significant challenges in reliably extracting physiological signals, especially in real-world scenarios.
Recent advances in deep learning have leveraged convolutional neural network (CNN) and vision transformer (ViT)~\citep{dosovitskiy2020image} for rPPG estimation, owing to their ability to learn representations of periodic signals from facial video.
These methods have demonstrated superior accuracy over traditional approaches, as they can effectively extract rPPG signals under subject motion and illumination variations.
However, collecting large-scale labeled rPPG datasets is inherently challenging due to the need for synchronized physiological signals and facial video recordings.
Consequently, fully supervised deep learning models demonstrate strong intra-dataset performance but limited cross-dataset generalization due to distribution shifts.

Self-supervised learning (SSL) has been developed to address the limitations of fully supervised methods, enabling models to learn scalable and generic features from unlabeled data.
Among SSL approaches, variational auto-encoder~\citep{kingma2013auto}, contrastive learning~\citep{chen2020simple} and self-supervised ViT have shown significant effectiveness in visual representation learning.
In particular, masked autoencoder (MAE)~\citep{he2022masked} has proven beneficial for various downstream tasks by learning robust representations through pretext tasks.
The MAE framework masks a portion of the input image patches and trains a ViT to reconstruct the missing regions, thereby encouraging the encoder to capture meaningful input representations.  
Beyond general computer vision tasks, recent rPPG studies have explored SSL, and reported promising results~\citep{sun2024contrast, yue2023facial, shao2023tranphys, liu2024rppg}.
However, further improvements are still required to achieve robust and accurate rPPG estimation, particularly under cross-dataset evaluation settings.

To achieve reliable rPPG estimation across diverse environments and subjects, enhancing the representational capacity of the model is essential.
While contrastive learning does not require labeled data, it often incurs high computational costs.  
Furthermore, contrastive learning primarily focuses on relationships between samples rather than intrinsic data properties, limiting its effectiveness for robust rPPG representation learning.  
Instead, several recent studies~\citep{shao2023tranphys, liu2024rppg} have explored MAE-based pre-training for the rPPG domain.  
Liu et al.~\citep{liu2024rppg} proposed rPPG-MAE, which pre-trains a vanilla ViT encoder using spatio-temporal maps (STMaps)~\citep{niu2019rhythmnet} and subsequently fine-tunes it for rPPG signal regression.  
However, rPPG-MAE operates on STMaps as input, requiring additional preprocessing steps that may discard fine-grained spatio-temporal information and adversely affect physiological signal estimation performance.  
In contrast to previous works, we focus on learning rPPG-specific representations directly from raw facial videos in a self-supervised manner.  
Rather than adapting MAE as a generic pre-training tool, we redesign the MAE framework to explicitly account for the periodic and physiological characteristics of rPPG signals.  
Specifically, our method adopts the VideoMAE training paradigm~\citep{tong2022videomae} while incorporating rPPG-oriented inductive biases, including a periodic masking strategy and frequency-domain constraints.  
These domain-specific modifications enable our model to learn physiologically meaningful periodic representations that encode both temporal dynamics and frequency characteristics during pre-training, thereby facilitating robust rPPG estimation under diverse and unseen data distributions.

\begin{figure}[t]
  \centering
   \includegraphics[width=0.75\linewidth]{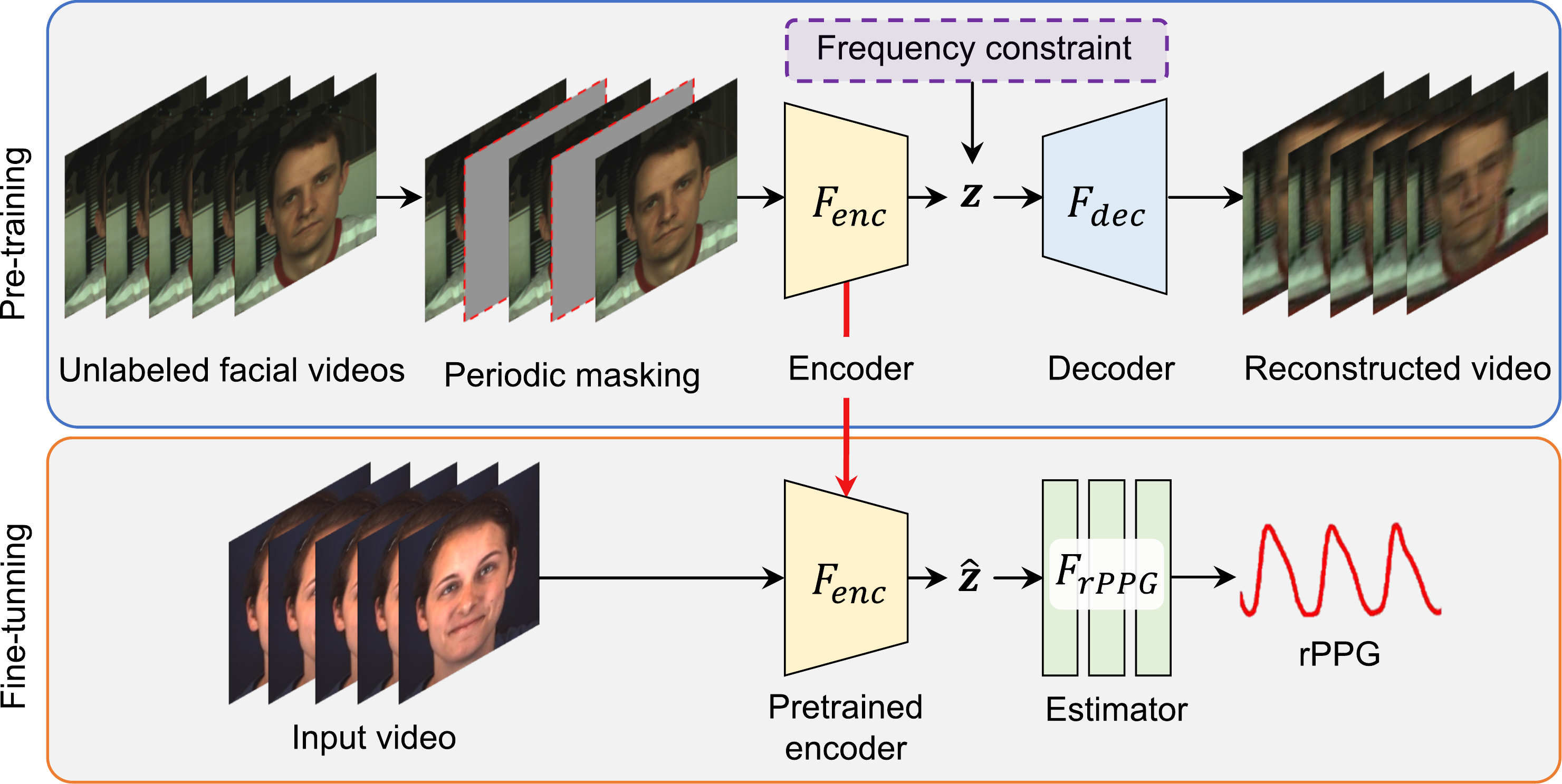}
   \caption{Overview of the proposed framework. We pre-train the model for the downstream task of rPPG estimation in a self-supervised manner using the periodic masking strategy. Additionally, the pre-training model is constrained by the loss function defined in the frequency domain to capture the desired physiological signals.}
   \label{fig:overview}
\end{figure}

Conventional masking methods proposed in VideoMAE~\citep{tong2022videomae} do not consider periodicity and are therefore unsuitable for the rPPG domain.
To address this limitation, we propose a novel masking strategy that periodically masks input frames based on a resampling approach.
This strategy encourages the encoder to learn diverse periodic patterns present in the masked input frames, enabling it to capture the inherent periodic nature of physiological signals.
Additionally, we integrate frequency constraints into the MAE framework to enhance the learning of desired physiological features within spatio-temporal latent representations, enabling the model to effectively capture rhythmic patterns in unlabeled facial videos.
Figure~\ref{fig:overview} provides an overview of the proposed framework, and the main contributions are as follows:

\begin{itemize}
\item We propose Periodic-MAE, a rPPG data-agnostic self-supervised framework that learns generic and transferable representations directly from unlabeled facial videos.
\item We introduce the periodic masking strategy, which enhances the learning of inherent periodic features and facilitates rPPG-oriented representation learning
\item We incorporate physiological frequency constraints into the MAE pre-training framework to encourage the model to focus on physiologically relevant cues during reconstruction.
\item Extensive experiments on public benchmarks and a real-world dataset demonstrate that our method learns robust representations with strong cross-dataset generalization.

\end{itemize}

\section{Related work}
\label{sec:related work}

\subsection{rPPG estimation}
rPPG is a camera-based method for measuring vital signals in a non-contact manner.
Conventional rPPG approaches~\citep{balakrishnan2013detecting, poh2010non, de2013robust, wang2016algorithmic} have relied on signal processing to detect periodic signals in video using hand-crafted features.
With the advancement of data-driven deep learning techniques, CNN and transformer architectures have been widely applied to video analysis tasks.
In particular, PhysNet~\citep{yu2019remote} has demonstrated promising performance in heart rate estimation using a three-dimensional CNN (3D CNN), which effectively captures temporal information.
Although many studies have utilized 3D CNN, convolution-based operations primarily learn spatial and temporal correlations between neighboring frames and are limited in capturing long-range dependencies.

ViT can learn long-range correlations across spatial and temporal dimensions, which facilitates video analysis and the estimation of quasi-periodic signals such as rPPG.
PhysFormer~\citep{yu2022physformer} and PhysFormer$++$~\citep{yu2023physformer++} introduced a video transformer that extracts long-term spatio-temporal features and incorporates an attention module that leverages temporal differences.
Zou et al.~\citep{zou2025rhythmformer} proposed RhythmFormer, a transformer-based architecture with a sparse attention mechanism tailored for periodic signals.
However, most transformer-based studies~\citep{yu2022physformer, yu2023physformer++, zou2025rhythmformer} rely on supervised learning and are constrained by the limited dataset size, which impedes the development of generalizable representations and leads to suboptimal performance, especially in cross-dataset evaluations.

\subsection{Self-supervised learning in rPPG}
Recently, contrastive learning and MAE have been adopted for estimating remote physiological signals.
Yue et al.~\citep{yue2023facial} trained an rPPG estimation model in a self-supervised manner using positive and negative samples.
They generated negative samples by modulating the frequency and defined the spatially augmented clips as positive samples.
CalibrationPhys~\citep{akamatsu2023calibrationphys} utilized data from several cameras to train the model on unlabeled video, leveraging different views of the same subject to attract the predicted pulse and repel signals from other individuals.
However, contrastive learning techniques are not data-agnostic as their pairs are determined within the same dataset, and therefore fail to learn generic representations.

MAE has demonstrated its effectiveness in downstream tasks by leveraging rich representations in a self-supervised manner and has also been applied to rPPG estimation.
The first attempt was rPPG-MAE, proposed by \citep{liu2024rppg}, to model the self-similarity of physiological signals by reconstructing STMap.
However, they adopted a two-dimensional MAE and generated various types of STMap as input instead of using the facial video directly.
TranPhys~\citep{shao2023tranphys} utilizes the VideoMAE~\citep{tong2022videomae} and tube masking strategy to reconstruct the YUV map from masked tokens of the input video.
The previous method requires additional preprocessing to optimize the reconstruction pipeline, which may result in the loss of fine-grained information and increased susceptibility to noise, potentially degrading the performance of deep learning models depending on the region of interest selection.
In contrast, the proposed method directly reconstructs RGB videos and incorporates physiological frequency-domain constraints into the pre-training pipeline to guide the model in learning general representations of the rPPG signal.

\section{Proposed method}
\begin{figure*}[t]
  \centering
   \includegraphics[width=1.0\linewidth]{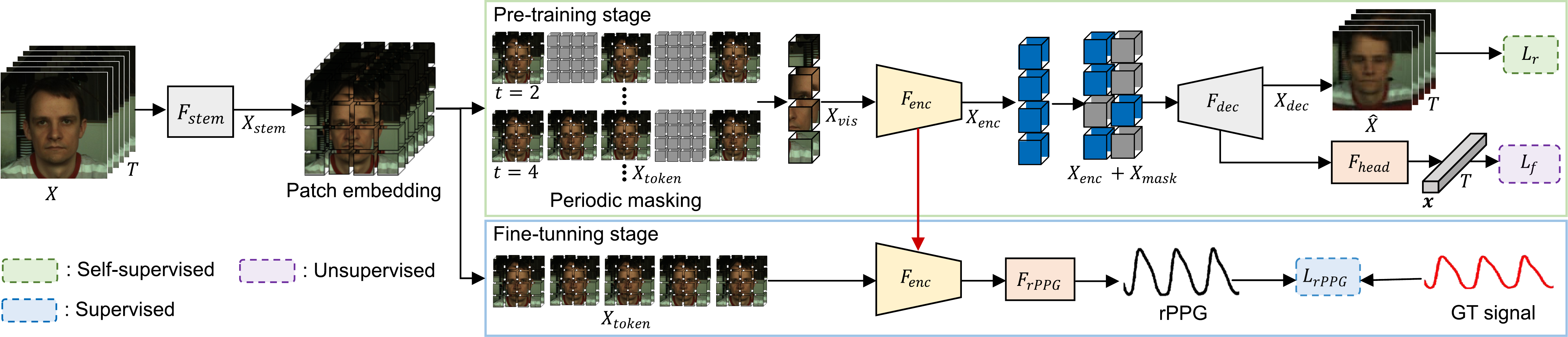}
   \caption{Overview of the Periodic-MAE. Our framework consists of pre-training and fine-tuning stages, depicted by blue and red arrows, respectively. The black arrow represents the pipeline for both stages. In the pre-training stage, we leverage a ViT-based encoder and decoder with a periodic masking strategy to reconstruct periodically missing frames. Additionally, frequency-domain constraints based on the physiological signal nature are applied to ensure the model focuses on the pulse signal.} 
   \label{fig2}
\end{figure*}

rPPG estimation is based on the skin epidermal and subcutaneous physiological priors with the facial region and the subtle changes in skin tone.
Our objective is to learn robust and transferable universal periodic signal cues and facial region representation from unlabeled datasets.
To this end, we employ a pretext task that reconstructs masked regions using an asymmetric encoder-decoder architecture.
The overall framework of the proposed framework is shown in Figure~\ref{fig2}.

Given an input video $X\in \mathbb{R}^{T\times H \times W \times 3}$, where $T$, $H$, and $W$ denote the number of frames, height, and width, respectively, we first employ a fusion stem module $F_{stem}$ inspired by~\citep{zou2025rhythmformer}.
This module fuses the raw frames with their corresponding frame differences to guide the transformer's self-attention toward capturing subtle skin tone changes rather than more prominent facial features, thereby enhancing rPPG signal extraction.
The stem module outputs $X_{stem} = F_{stem}(X)$, with $X_{stem} \in \mathbb{R}^{T\times \frac{H}{4} \times \frac{W}{4} \times C}$, where $C$ is the number of channels.
Next, $X_{stem}$ is partitioned into non-overlapping spatio-temporal tokens $X_{token} \in \mathbb{R}^{T\times \frac{H}{16} \times \frac{H}{16} \times C}$ via a patch embedding process.
This tokenization aggregates neighboring semantic information and reduces the total number of tokens to $T\times \frac{H}{16}\times \frac{W}{16}$, significantly lowering the computational cost of subsequent transformer operations. 

\begin{figure}[t]
  \centering
   \includegraphics[width=1.0\linewidth]{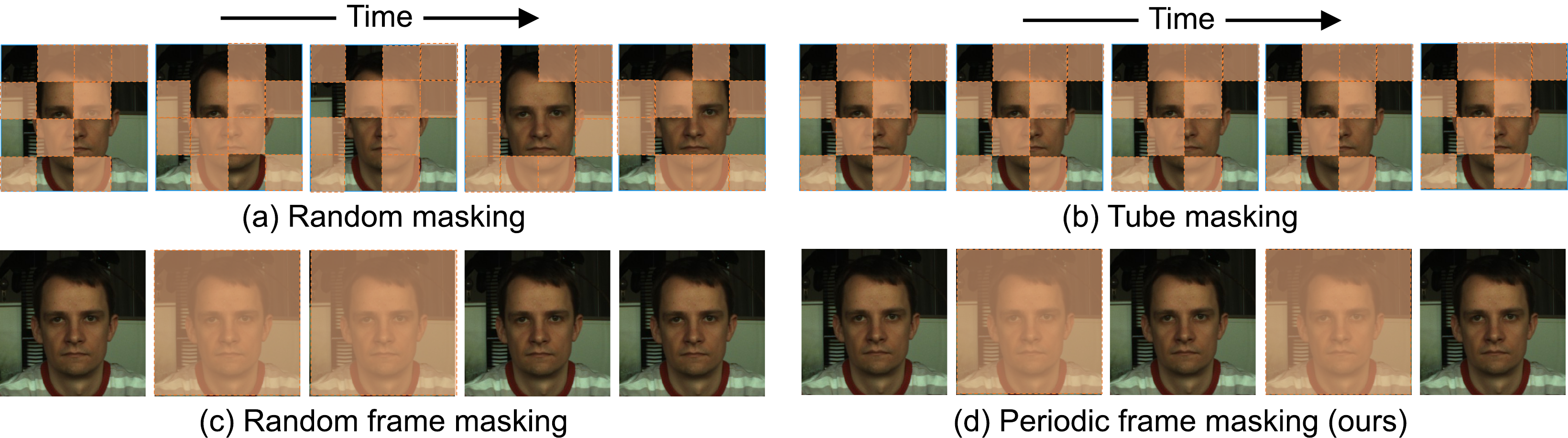}
   \caption{Comparison of masking strategies. (a) random masking, (b) tube masking, (c) random frame masking, and (d) proposed periodic masking that samples frames at regular intervals with step size $t\in [2,5]$, randomly varied across epochs to preserve temporal periodicity while preventing trivial reconstruction.}
   \label{fig3}
\end{figure}

\subsection{Self-supervised representation learning for rPPG}
\textbf{Periodic frame masking.} We propose a novel masking strategy based on signal periodicity, as shown in Figure~\ref{fig3} (b).
In the rPPG task, data augmentation typically involves adjusting the number of frames and resampling the signal to align with the adjusted video length~\citep{hwang2024phase, yu2020autohr, lokendra2022and}, thereby generating diverse cardiac pulse signals.
Building on this approach, we introduce the periodic frame masking approach that explicitly exploits the rhythmic nature of physiological signals.
This strategy enables the model to reconstruct periodically missing signals using contextual information from neighboring visible features, enhancing its ability to learn diverse rhythmic patterns and improving the spatiotemporal representation of facial videos.

Conventional masking techniques, such as patch, tube, and frame masking, reduce redundant information from input frames, facilitating robust representation learning~\citep{tong2022videomae, shao2023tranphys}.
However, these approaches do not explicitly leverage the periodic nature of physiological signals and retain a fixed number of tokens based on a predetermined masking ratio.
To address this, we introduce periodic masking, which selects frames at regular intervals with a step size $t \in [2,5]$, randomly varying $t$ across epochs to simulate different downsampling effects and prevent trivial reconstruction.
This strategy preserves evenly spaced temporal information, helping the model learn the periodic patterns while reconstructing missing signals.
Consistent with existing MAE frameworks for rPPG tasks~\citep{tong2022videomae, shao2023tranphys}, we maintain a masking ratio of 50 \% and 80 \% within this step size range.
By enforcing reconstruction based on physiological priors, the model uncovers rhythmic patterns, extracting generic physiological features across diverse facial videos.
The effectiveness of the periodic masking method is validated through the ablation studies in Table~\ref{tab:4}.

\textbf{Video masked autoencoder.} After the periodic masking, the encoder $F_{enc}$ processes the unmasked visible tokens $X_{vis}$ to learn a representation of the physiological signal.
We adopt a ViT-based model specifically designed for rPPG~\citep{zou2025rhythmformer}.
The encoder output $X_{enc}$ consists of features corresponding to the visible tokens.
The decoder input consists of $X_{enc}$ and $X_{mask}$, which represent learnable masked tokens, and are used to reconstruct the missing frames.
The tokens combined with the positional embedding fed into the decoder $F_{dec}$ for frame reconstruction.
Following the approach of VideoMAE, we employ a basic ViT decoder with the depth of 4.
Through the MAE training process, the model learns generalized representations across diverse facial videos.

\textbf{Optimizing Periodic-MAE.} Since the input video contains various noise components, it becomes challenging to extract rhythmic cues in a self-supervised pre-training stage.
To address this issue, we optimize the encoder and decoder by integrating frequency range constraints into the MAE framework.
Physiological signals are characterized by a finite bandwidth and exhibit sparse power spectra.
Previous work~\citep{speth2023non} demonstrated that applying frequency-domain constraints on deep learning models can effectively isolate pulse signals in facial video data.

Specifically, we focus on the typical frequency range of cardiac pulses.
These pulses occur between 40 beats per minute (bpm) and 180 bpm, corresponding to a frequency range of approximately $0.\overline{66}$ Hz to 3 Hz.
The specified frequency band guides the model in identifying sparse, periodic components from the input video.
For loss calculation, we utilize a one-dimensional convolutional layer, $F_{head}$, which processes the decoder output $X_{dec}$ to produce a signal $\textbf{x}$ of length $T$.
We then impose bandlimit constraints on the reconstructed signal by applying three distinct loss terms to $\textbf{x}$.
This penalty ensures that the frequency component of the output remains within a physiologically plausible range, thereby encouraging the model to reconstruct signals that accurately represent the underlying pulse signals.
The physiological prior further guides the reconstruction model in learning the relevant pulse signals from the input video.

The irrelevant power ratio is incorporated into the bandwidth loss $L_b$, penalizing the model for detecting signals outside the specified bandwidth, as follows.
\begin{equation}
     L_b = \frac{1}{\sum\limits_{i=-\infty}^{\infty} f_i}\left ( \sum_{i=-\infty}^{a}f_i +  \sum_{b}^{\infty}f_i \right ),
\end{equation}
\noindent
where $a$ and $b$ are set to $0.\overline{66}$ Hz and 3 Hz, respectively.
$f_i = FFT(\textbf{x})$ represents the power in the $i$-th frequency bin and $FFT$ is the fast Fourier transform.

The heartbeats result in a regular pulse, which appears as a distinct power component in the frequency domain. 
Therefore, we can effectively extract the desired signal by ignoring noise signals with weak periodicity and avoiding wide-band prediction.
We penalize energy within bandlimits that are distant from the spectral peak as follows.
\begin{equation}
     L_s = \frac{1}{\sum\limits_{i=a}^{b}f_i} \left ( \sum_{i=a}^{f^*-\Delta_f}f_i +  \sum_{f^*+\Delta_f}^{b}f_i \right ),
\end{equation}
\noindent
where, $f^*$ and $\Delta_f$ represent the spectral peak frequency and its surrounding margin, respectively.
Following~\citep{speth2023non}, we set $\Delta_f$ to 6 beats per minute.
We define the frequency-domain constraint loss as $L_f=L_b+L_s$.

For the reconstruction task, the mean squared error (MSE) is computed between the normalized masked tokens and reconstructed tokens.
\begin{equation}
    L_r = \frac{1}{\Omega}\sum_{p\in\Omega}\left \| X(p)-\hat{X}(p)\right \|_2,
\end{equation}
where $p$ denotes the token index, $\Omega$ is the set of all tokens, and $X$ and $\hat{X}$ represents the input and reconstructed videos, respectively.
The total loss for optimizing the pre-training stage is $L_{MAE} = L_f+L_r$.

\subsection{rPPG estimation}
In the pre-training stage, the model learns a generic rhythm representation from raw facial videos in a self-supervised manner.
For the downstream task of rPPG estimation, $F_{enc}$ is fine-tuned and $F_{rppg}$ predicts the rPPG signal.
The $F_{rppg}$ is constructed with linear layers.
We input full images without masking into the model, which differs from the reconstruction task.
The output is a vector of length 160, matching the length of the input video.
To train the rPPG estimator, we employ three loss functions in both the time and frequency domains.

In the time domain, the negative Pearson correlation loss is calculated between the estimated rPPG and ground truth signal.
\begin{equation}
   L_{time} =  1 - \rho(x,x'),
\end{equation}
where $\rho(x,x')$ is the Pearson correlation coefficient between the predicted rPPG signals $x$ and the ground truth cardiac pulse $x'$.
This loss function ensures the predicted signal has a similar trend and peak values to the sensor signal on the time axis.

We calculate the power spectral density (PSD) using a fast Fourier transform into the estimated and the sensor signals.
The frequency analysis can reveal the periodicity of the physiological signals which is collected while the heart beats~\citep{hwang2024phase}.
The loss function defined in the frequency domain is as follows.
\begin{equation}
    L_{freq} = \left \| P(x)-P(x') \right \|_2,
\end{equation}
where $P(\cdot)$ is PSD.
$L_{freq}$ improves the robustness to noise by aligning the rPPG more closely with the ground truth and minimizing the errors of frequency components.
The overall loss for the rPPG estimation is defined as $L_{rPPG}=  L_{time} + \lambda L_{freq}$, where the hyperparameter $\lambda$ is set to 0.5.

\begin{figure}[t]
  \centering
   \includegraphics[width=0.7\linewidth]{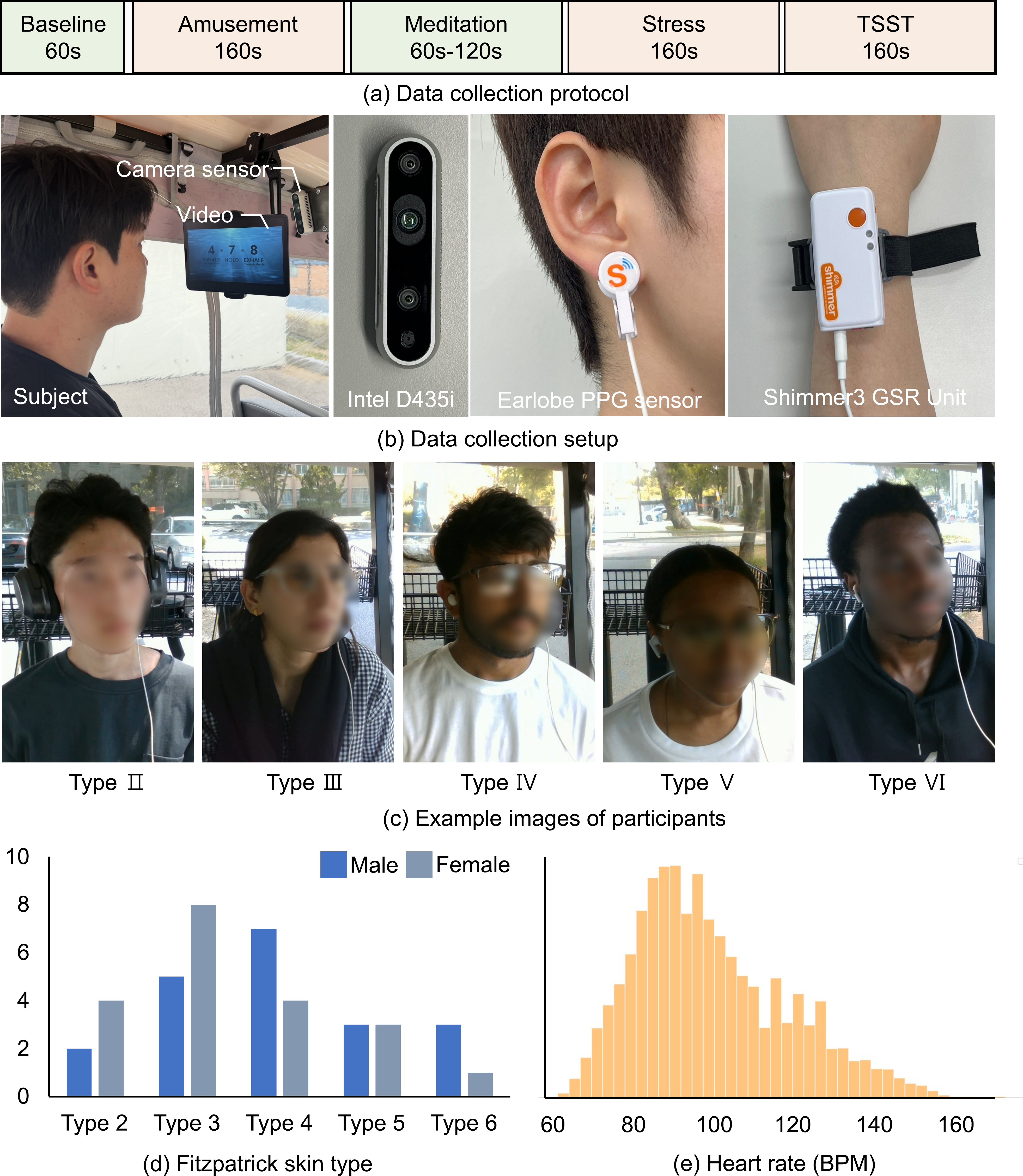}
   \caption{Overview of the RPED dataset.}
   \label{fig5}
\end{figure}

\section{Experiments}
\subsection{Datasets}
We use the pulse rate detection dataset (PURE)~\citep{stricker2014non}, UBFC-rPPG~\citep{bobbia2019unsupervised}, and the multi-domain mobile video physiology dataset (MMPD)~\citep{tang2023mmpd} for both self-supervised pre-training and the downstream rPPG estimation task.  
The Vision for Vitals (V4V) dataset~\citep{revanur2021first} and the collected real-world remote PPG and emotion detection (RPED) dataset are used exclusively for rPPG estimation.
For the PURE, UBFC-rPPG, and MMPD datasets, we follow the data split protocol proposed in~\cite{zou2025rhythmformer}, and the same splits are consistently used for both pre-training and intra-dataset testing.
The data split for V4V follows the protocol described in~\cite{hwang2024phase}.

\textbf{PURE} consists of recordings from 10 participants across 6 different activities. 
Each video is 1 minute long, with the resolution of 640$\times$480 and the frame rate of 30 Hz.
The 6 activities include steady, talking, slow head translation, fast head translation, and small and medium head rotations.
The training set comprises the first 60 \% of samples, while the remaining 40 \% is used for testing.

\textbf{UBFC-rPPG} includes 42 videos, each lasting 1 minute with the resolution of 640$\times$480 and the frame rate of 30 Hz.
42 participants performed mathematical tasks to induce variations in heart rates while remaining stationary state.
We use 30 subjects for training and the remaining 12 subjects for testing.

\textbf{MMPD} contains videos from 33 subjects, totaling 11 hours of video recorded on mobile phones. 
It consists of 660 videos, each 1-minute long with the resolution of 320$\times$240 and the frame rate of 30 Hz.
This dataset encompasses a various skin tone, motion, and lighting conditions.
Specifically, the body motion is comprised of static, head motion, talking and walking under LED-high and low, incandescent, and natural light.
The MMPD is available in both compressed and uncompressed versions.
We used the compressed version for experimentation and divided it into 70 \% for training, 10 \% for validation, and 20 \% for testing.

\textbf{V4V} contains a total of 1,358 videos collected from 179 subjects, with the resolution of 1280$\times$720 and the frame rate of 25 fps.
Each video varies in length and includes 10 activities designed to capture a range of emotional states.
The dataset is split into 724 samples for training, 276 for validation, and 358 for testing.

\textbf{RPED dataset} comprises video recordings of 40 subjects captured in unconstrained environments.
Participants were seated while watching videos designed to induce emotional responses across phases: meditation, amusement stimuli, and stress induction, followed by the Trier Social Stress Test (TSST), as shown in Figure~\ref{fig5}.
Each recording session lasted approximately 10 minutes and was captured with the resolution of 1920$\times$1080 and a frame rate of 30 Hz.
Ground truth PPG signals were recorded using a Shimmer3R GSR$+$ Unit earlobe sensor at 128 Hz.

During acquisition, participants were allowed to move naturally within the camera's field of view. All recordings were conducted exclusively under ambient lighting without supplementary artificial illumination, resulting in illumination variability both within and across sessions. This variability reflects realistic operational conditions, thereby enabling robust evaluation of rPPG algorithms under real-world deployment scenarios.
To ensure demographic diversity, the 40 subjects were recruited to represent a broad spectrum of Fitzpatrick skin types, as illustrated in Figure~\ref{fig5}. For experimental evaluation, the dataset was partitioned into 80 \% for training and 20 \% for testing.
All data collection procedures were conducted under the approval of the Institutional Review Board (IRB Number: JBNU 2025-04-035-002).

\subsection{Implementation details and evaluation metric}
\textbf{Preprocessing}. We implemented the proposed method using the rPPG-toolbox~\citep{liu2024rppg}.
The face region was cropped and resized to 128$\times$128 as input for the pre-training and fine-tuning stages.
The first frame was selected and fixed for the subsequent frames, and we utilized 160 frames for training.

\textbf{Pre-training}. We used the PURE, UBFC-rPPG, and MMPD datasets to train the model on the reconstruction task.
The encoder and decoder were pre-trained with each training set from these datasets.
We used the AdamW optimizer with momentum parameters $\beta_1$ is 0.9, $\beta_2$ is 0.95, weight decay of 0.05, and a base learning rate of 0.1.
The learning rate was scheduled using cosine decay with a linear scaling rule.
The batch size was set to 4 and was trained for 120 Epochs.
Our model was implemented in PyTorch and deployed on NVIDIA RTX 6000 Ada.

\textbf{rPPG estimation}. We fine-tuned the encoder from the pre-training stage and the rPPG estimator using PPG labels.
The AdamW optimizer was utilized with a batch size of 8 and a learning rate of 0.001.
The model was trained for 30 epochs, with the random seed fixed to ensure reproducibility.
The proposed model was implemented in PyTorch and deployed on NVIDIA RTX 6000 Ada.

\textbf{Evaluation metric}. We report five metrics to evaluate the HR estimation task, which include mean absolute error (MAE), root mean squared error (RMSE), mean absolute percentage error (MAPE), Pearson correlation coefficient $\rho$, and signal-to-noise ratio (SNR).	

\begin{table*}[t]
\centering
\addtolength{\tabcolsep}{-0.3em}
\renewcommand{\arraystretch}{1.2}
\resizebox{1.0\linewidth}{!}{%
\begin{tabular}{l|c|ccccccccccccccc}
\hline
\multirow{3}{*}{Method}                    & \multicolumn{1}{l|}{\multirow{3}{*}{Train set}} & \multicolumn{15}{c}{Test set}                                                                                                                                                                                                                                                               \\ \cline{3-17} 
                                           & \multicolumn{1}{l|}{}                           & \multicolumn{5}{c|}{PURE}                                                                           & \multicolumn{5}{c|}{MMPD}                                                                            & \multicolumn{5}{c}{V4V}                                                        \\ \cline{3-17} 
                                           & \multicolumn{1}{l|}{}                           & MAE           & RMSE          & MAPE          & $\rho$        & \multicolumn{1}{c|}{SNR}            & MAE           & RMSE           & MAPE          & $\rho$        & \multicolumn{1}{c|}{SNR}            & MAE           & RMSE          & MAPE          & $\rho$        & SNR            \\ \hline
DeepPhys~\citep{chen2018deepphys}          & \multirow{7}{*}{UBFC}                           & 5.54          & 18.51         & 5.32          & 0.66          & \multicolumn{1}{c|}{4.40}           & 17.50         & 25.00          & 19.27         & 0.06          & \multicolumn{1}{c|}{-11.72}         & 8.21          & 14.99         & 9.52          & 0.46          & -4.84          \\
PhysNet~\citep{yu2019remote}               &                                                 & 8.06          & 19.71         & 13.67         & 0.61          & \multicolumn{1}{c|}{6.68}           & 9.47          & 16.01          & 11.11         & 0.31          & \multicolumn{1}{c|}{-8.15}          & 5.49          & 9.46          & 9.85          & 0.47          & -4.17          \\
TS-CAN~\citep{liu2020multi}                &                                                 & 3.69          & 13.8          & 3.39          & 0.82          & \multicolumn{1}{c|}{5.26}           & 14.01         & 21.04          & 15.48         & 0.24          & \multicolumn{1}{c|}{-10.18}         & 5.53          & 10.98         & 6.35          & 0.63          & -4.52          \\
PhysFormer~\citep{yu2022physformer}        &                                                 & 2.93          & 10.11         & 4.80          & 0.90          & \multicolumn{1}{c|}{9.91}           & 10.69         & 17.23          & 12.48         & 0.27          & \multicolumn{1}{c|}{-8.84}          & 6.62          & 11.39         & 7.52          & 0.50          & -3.96          \\
EfficientPhys~\citep{liu2023efficientphys} &                                                 & 5.47          & 17.04         & 5.40          & 0.71          & \multicolumn{1}{c|}{4.09}           & 13.78         & 22.25          & 15.15         & 0.09          & \multicolumn{1}{c|}{-9.13}          & 5.19          & 10.34         & 5.96          & 0.67          & -3.89          \\
RhythmFormer~\citep{zou2025rhythmformer}   &                                                 & 0.97          & 3.36          & 1.60          & 0.99          & \multicolumn{1}{c|}{\textbf{12.01}} & 9.08          & 15.07          & 11.17         & 0.53 & \multicolumn{1}{c|}{-7.73}          & 6.92          & 13.56         & 8.08          & 0.56 & -2.73          \\
Periodic-MAE (ours)                        &                                                 & \textbf{0.75} & \textbf{2.42} & \textbf{1.22} & \textbf{0.99} & \multicolumn{1}{c|}{8.69}           & \textbf{7.85} & \textbf{13.97} & \textbf{8.95} & \textbf{0.53} & \multicolumn{1}{c|}{\textbf{-6.55}} & \textbf{4.48} & \textbf{9.35} & \textbf{5.25} & \textbf{0.69} & \textbf{-1.67} \\ \hline
\end{tabular}%
}
\caption{Cross-dataset evaluation results for HR estimation. The model is trained on UBFC-rPPG and tested on PURE, MMPD, and V4V. The best results are highlighted in bold.}
\label{tab:cross-data-1}
\end{table*}			

\begin{table*}[t]
\centering
\addtolength{\tabcolsep}{-0.3em}
\resizebox{1.0\linewidth}{!}{%
\begin{tabular}{l|c|ccccccccccccccc}
\hline
\multirow{3}{*}{Method}                    & \multicolumn{1}{l|}{\multirow{3}{*}{Train set}} & \multicolumn{15}{c}{Test set}                                                                                                                                                                                                                                                                \\ \cline{3-17} 
                                           & \multicolumn{1}{l|}{}                           & \multicolumn{5}{c|}{UBFC}                                                                          & \multicolumn{5}{c|}{MMPD}                                                                             & \multicolumn{5}{c}{V4V}                                                         \\ \cline{3-17} 
                                           & \multicolumn{1}{l|}{}                           & MAE           & RMSE          & MAPE          & $\rho$        & \multicolumn{1}{c|}{SNR}           & MAE           & RMSE           & MAPE           & $\rho$        & \multicolumn{1}{c|}{SNR}            & MAE           & RMSE           & MAPE          & $\rho$        & SNR            \\ \hline
DeepPhys~\citep{chen2018deepphys}          & \multirow{7}{*}{PURE}                           & 1.21          & 2.90          & 1.42          & 0.99          & \multicolumn{1}{c|}{1.74}          & 16.92         & 24.61          & 18.54          & 0.05          & \multicolumn{1}{c|}{-11.53}         & 6.33          & 12.56          & 7.28          & 0.57          & -4.01          \\
PhysNet~\citep{yu2019remote}               &                                                 & 0.98          & 2.48          & 1.12          & 0.99          & \multicolumn{1}{c|}{1.49}          & 13.94         & 21.61          & 15.15          & 0.20          & \multicolumn{1}{c|}{-9.94}          & 4.22          & 8.91           & 5.18          & 0.76          & -4.63          \\
TS-CAN~\citep{liu2020multi}                &                                                 & 1.30          & 2.87          & 1.50          & 0.99          & \multicolumn{1}{c|}{1.49}          & 13.94         & 21.61          & 15.15          & 0.20          & \multicolumn{1}{c|}{-9.94}          & 5.35          & 10.90          & 6.23          & 0.63          & -3.85          \\
PhysFormer~\citep{yu2022physformer}        &                                                 & 2.35          & 6.26          & 2.32          & 0.94          & \multicolumn{1}{c|}{2.97}          & 10.49         & 17.14          & 11.98          & 0.32          & \multicolumn{1}{c|}{-8.27}          & 6.99          & 13.97          & 8.51          & 0.57          & -5.17          \\
EfficientPhys~\citep{liu2023efficientphys} &                                                 & 2.07          & 6.32          & 2.10          & 0.94          & \multicolumn{1}{c|}{-0.12}         & 14.03         & 21.62          & 15.32          & 0.17 & \multicolumn{1}{c|}{-9.95}          & 6.33          & 12.56          & 7.28          & 0.57 & -4.01          \\
RhythmFormer~\citep{zou2025rhythmformer}   &                                                 & 0.89 & 1.83 & 0.97 & 0.99 & \multicolumn{1}{c|}{6.05} & 8.98 & 14.85 & 11.11 & \textbf{0.51} & \multicolumn{1}{c|}{-8.39} & 5.11 & 11.30 & 5.83 & 0.65 & -0.77 \\
Periodic-MAE (ours)                        &                                                 & \textbf{0.86} & \textbf{1.78} & \textbf{0.95} & \textbf{0.99} & \multicolumn{1}{c|}{\textbf{7.84}} & \textbf{8.33} & \textbf{13.60} & \textbf{9.29}  & 0.47 & \multicolumn{1}{c|}{\textbf{-8.12}} & \textbf{2.94} & \textbf{6.95}  & \textbf{3.45} & \textbf{0.83} & \textbf{-0.57} \\ \hline
\end{tabular}%
}
\caption{Cross-dataset evaluation results for HR estimation. The model is trained on PURE and tested on UBFC-rPPG, MMPD, and V4V. The best results are highlighted in bold.}
\label{tab:cross-data-2}
\end{table*}

\begin{figure*}[t]
  \centering
   \includegraphics[width=1.0\linewidth]{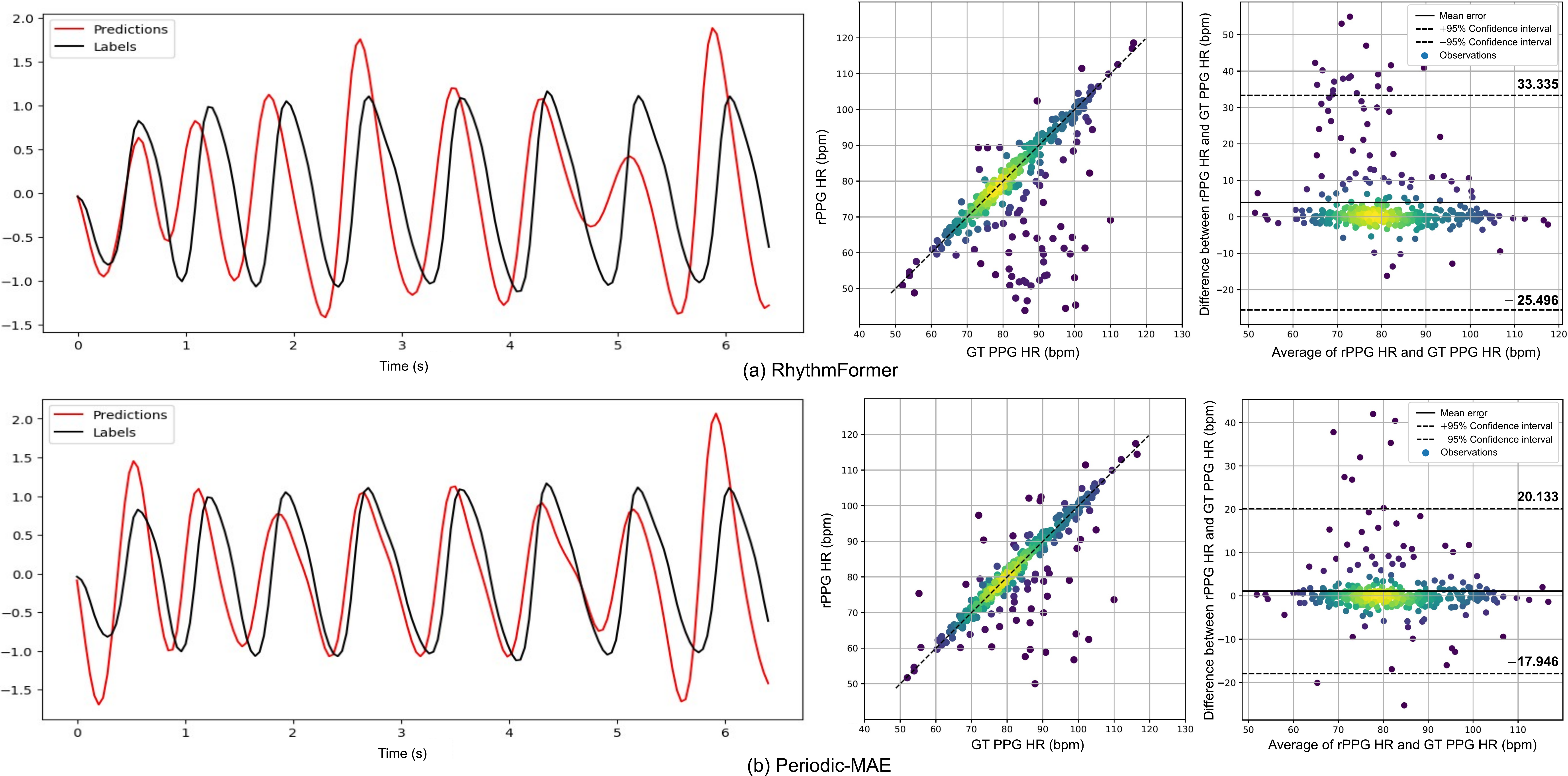}
   \caption{Visualization of PPG and rPPG signals, along with Bland-Altman plots for cross-dataset testing results on V4V using a model trained on PURE.}
   \label{fig4}
\end{figure*}

\begin{table*}[t]
\centering
\addtolength{\tabcolsep}{-0.3em}
\renewcommand{\arraystretch}{1.2}
\resizebox{0.8\linewidth}{!}{%
\begin{tabular}{l|cccccccccc}
\hline
\multirow{3}{*}{Method}                    & \multicolumn{10}{c}{Test set}                                                                                                                                                                 \\ \cline{2-11} 
                                           & \multicolumn{5}{c|}{UBFC $\rightarrow$ RPED}                                                                  & \multicolumn{5}{c}{PURE $\rightarrow$ RPED}                                               \\ \cline{2-11} 
                                           & MAE            & RMSE           & MAPE           & $\rho$        & \multicolumn{1}{c|}{SNR}             & MAE            & RMSE           & MAPE           & $\rho$         & SNR             \\ \hline
DeepPhys~\citep{chen2018deepphys}          & 21.07          & 26.53          & 23.81          & -0.01         & \multicolumn{1}{c|}{-15.05}          & 21.10          & 27.32          & 23.58          & -0.07          & -14.47          \\
PhysNet~\citep{yu2019remote}               & 16.75          & 24.85          & 20.94          & 0.13          & \multicolumn{1}{c|}{-11.97}          & 37.19          & 45.62          & 45.32          & 0.05           & -15.75          \\
TS-CAN~\citep{liu2020multi}                & 20.54          & 26.94          & 23.83          & -0.02         & \multicolumn{1}{c|}{-13.52}          & 21.27          & 27.69          & 24.47          & -0.01          & -13.71          \\
PhysFormer~\citep{yu2022physformer}        & 15.71          & 23.09          & 19.32          & 0.01          & \multicolumn{1}{c|}{-11.38}          & 28.28          & 37.36          & 34.59          & -0.03          & -14.29          \\
EfficientPhys~\citep{liu2023efficientphys} & 17.78          & 24.97          & 20.08          & 0.01          & \multicolumn{1}{c|}{-11.60}          & 18.82          & 25.16          & 21.34          & -0.04 & -13.17          \\
RhythmFormer~\citep{zou2025rhythmformer}   & 13.88 & 19.33 & 16.89 & 0.01 & \multicolumn{1}{c|}{\textbf{-11.75}} & 32.79 & 38.99 & 39.86 & -0.03 & -17.08 \\
Periodic-MAE (ours)                        & \textbf{12.75} & \textbf{17.56} & \textbf{15.24} & \textbf{0.14} & \multicolumn{1}{c|}{-11.73} & \textbf{16.75} & \textbf{21.78} & \textbf{19.07} & \textbf{0.08}  & \textbf{-14.03} \\ \hline
\end{tabular}%
}
\caption{Experimental results of the cross-dataset evaluation for HR estimation on the RPED dataset. The best results are marked in bold.}
\label{tab:cross-data-3}
\end{table*}

\begin{table*}[t]
\centering
{\scriptsize
\addtolength{\tabcolsep}{-0.15em}
\renewcommand{\arraystretch}{1.2}
\resizebox{1.0\linewidth}{!}{%
\begin{tabular}{l|ccc|ccc|ccc|ccc}
\hline
\multirow{2}{*}{Method}                    & \multicolumn{3}{c|}{PURE}                     & \multicolumn{3}{c|}{UBFC-rPPG}                & \multicolumn{3}{c|}{MMPD} & \multicolumn{3}{c}{V4V} \\ \cline{2-13} 
                                           & MAE           & RMSE          & $\rho$        & MAE           & RMSE          & $\rho$        & MAE    & RMSE   & $\rho$  & MAE    & RMSE  & $\rho$ \\ \hline
DeepPhys~\citep{chen2018deepphys}          & 0.83          & 1.54          & 0.99          & 6.27          & 10.82         & 0.65          & 22.27  & 28.92  & -0.03   & 10.20  & 13.25 & 0.45   \\
PhysNet~\citep{yu2019remote}               & 2.10          & 2.6           & 0.99          & 2.95          & 3.67          & 0.97          & 4.80   & 11.80  & 0.60    & 13.15  & 19.23 & 0.75   \\
TS-CAN~\citep{liu2020multi}                & 2.48          & 9.01          & 0.92          & 1.70          & 2.72          & 0.99          & 9.71   & 17.22  & 0.44    & -      & -     & -      \\
Siamese~\citep{tsou2020siamese}            & 0.51          & 1.56          & 0.83          & 0.48          & 0.97          & -             & -      & -      & -       & -      & -     & -      \\
Dual-GAN~\citep{lu2021dual}                & 0.82          & 1.31          & 0.99          & 0.44          & 0.67          & 0.99          & -      & -      & -       & -      & -     & -      \\
TDM~\citep{comas2022efficient}             & 1.83          & 2.30          & 0.99          & 2.32          & 3.08          & 0.99          & -      & -      & -       & -      & -     & -      \\
PhysFormer~\citep{yu2022physformer}        & 1.10          & 1.75          & 0.99          & 0.50          & 0.71          & 0.99          & 11.99  & 18.41  & 0.18    & -      & -     & -      \\
EfficientPhys~\citep{liu2023efficientphys} & -             & -             & -             & 1.14          & 1.81          & 0.99          & 13.47  & 21.32  & 0.21    & -      & -     & -      \\
Li et al.~\citep{li2023learning}           & 1.44          & 2.50          & -             & 0.76          & 1.62          & -             & -      & -      & -       & -      & -     & -      \\
LSTC-rPPG~\citep{lee2023lstc}              & -             & -             & -             & 0.70          & 1.00          & 0.99          & -      & -      & -       & -      & -     & -      \\
Contrast-Phys+~\citep{sun2024contrast}     & 0.48          & 0.98          & 0.99          & \textbf{0.21}          & 0.80          & 0.99          & -      & -      & -       & -      & -     & -      \\
RhythmFormer~\citep{zou2025rhythmformer}   & 0.27          & 0.47          & 0.99          & 0.50          & 0.78          & 0.99          & 3.07   & 6.81   & 0.86    & -      & -     & -      \\
DRP-Net~\citep{hwang2024phase}             & -             & -             & -             & -             & -             & -             & -      & -      & -       & 3.83   & 9.59  & 0.75   \\
Periodic-MAE (ours)                        & \textbf{0.25} & \textbf{0.40} & \textbf{0.99} & 0.27 & \textbf{0.49} & \textbf{0.99} & \textbf{1.02}   & \textbf{3.78 }  & \textbf{0.96}    & \textbf{1.16}   & \textbf{3.32}  & \textbf{0.97}   \\ \hline
\end{tabular}%
}}
\caption{Comparative results of intra-dataset evaluation for HR estimation. The best results are marked in bold.}
\label{intra-data}
\end{table*}

\begin{table}[t]
\label{tab:mmdrive_intra}
\centering
\resizebox{0.6\linewidth}{!}{%
\renewcommand{\arraystretch}{1.2}
\begin{tabular}{l|ccccc}
\hline
Method              & MAE                  & RMSE                 & MAPE                 & $\rho$              & SNR                  \\ \hline
DeepPhys~\citep{chen2018deepphys}            & 19.257               & 23.286               & 24.814               & -0.041               & -15.093              \\
PhysNet~\citep{yu2019remote}             & 16.284               & 20.879               & 21.552               & -0.013               & -18.055              \\
TS-CAN~\citep{liu2020multi}             & 15.006                & 20.020                & 18.971                & 0.092     & -10.412\\
PhysFormer~\citep{yu2022physformer}          & 11.494               & 16.681               & 16.306               & 0.010                & -8.163               \\
EfficientPhys~\citep{liu2023efficientphys}       & 15.182               & 19.934               & 19.597               & 0.154                & -11.369              \\

RhythmFormer~\citep{zou2025rhythmformer}        & 11.256               & 15.668               & 16.078               & 0.080                & -9.457               \\
Periodic-MAE (ours) & \multicolumn{1}{c}{\textbf{8.440}} & \multicolumn{1}{c}{\textbf{13.258}} & \multicolumn{1}{c}{\textbf{12.258}} & \multicolumn{1}{c}{\textbf{0.415}} & \multicolumn{1}{c}{\textbf{-2.736}} \\ \hline
\end{tabular}%
}
\caption{Comparative results of intra-dataset evaluation for HR estimation on RPED dataset. The best results are marked in bold.}
\label{intra-data2}
\end{table}

\begin{table}[t]
\centering
\resizebox{0.6\linewidth}{!}{%
\renewcommand{\arraystretch}{1.2}
\begin{tabular}{l|ccc|ccc}
\hline
\multirow{2}{*}{Type} & \multicolumn{3}{c|}{PURE}                        & \multicolumn{3}{c}{UBFC-rPPG}                    \\ \cline{2-7} 
                      & MAE            & RMSE           & MAPE           & MAE            & RMSE           & MAPE           \\ \hline
Random                & 0.270          & 0.412          & 0.322          & 0.495          & 0.714          & 0.581          \\
Frame                 & 0.315          & 0.441          & 0.382          & 0.450          & 0.661          & 0.483          \\
Tube                  & 0.270          & 0.493          & \textbf{0.287} & 0.315          & 0.517          & 0.339          \\
Periodic (ours)       & \textbf{0.248} & \textbf{0.397} & 0.304          & \textbf{0.270} & \textbf{0.493} & \textbf{0.332} \\ \hline
\end{tabular}%
}
\caption{Ablation study on the masking strategy used during the pre-training stage. HR estimation results on the PURE and UBFC-rPPG datasets.}
\label{tab:4}
\end{table}

\begin{table}[t]
\centering
\resizebox{0.6\linewidth}{!}{%
\renewcommand{\arraystretch}{1.2}
\begin{tabular}{ccc|ccc|ccc}
\hline
\multirow{2}{*}{$L_r$} & \multirow{2}{*}{$L_b$} & \multirow{2}{*}{$L_s$} & \multicolumn{3}{c|}{PURE}                        & \multicolumn{3}{c}{UBFC-rPPG}                    \\ \cline{4-9} 
                       &                        &                        & MAE            & RMSE           & MAPE           & MAE            & RMSE           & MAPE           \\ \hline
\checkmark             &                        &                        & 0.293          & 0.427          & 0.364          & 0.450          & 0.731          & 0.536          \\
\checkmark             & \checkmark             &                        & 0.338          & 0.481          & 0.398          & 0.540          & 0.697          & 0.601          \\
\checkmark             &                        & \checkmark             & 0.293          & 0.427          & 0.339          & 0.495          & 0.810          & 0.614          \\
\checkmark             & \checkmark             & \checkmark             & \textbf{0.248} & \textbf{0.397} & \textbf{0.304} & \textbf{0.270} & \textbf{0.493} & \textbf{0.332} \\ \hline
\end{tabular}%
}
\caption{Ablation study on the loss functions used during the pre-training stage. HR estimation results on the PURE and UBFC-rPPG datasets.}
\label{tab:5}
\end{table}

\begin{table}[ht!]
\centering
\resizebox{0.5\columnwidth}{!}{%
\renewcommand{\arraystretch}{1.2}
\begin{tabular}{l|ccccc}
\hline
\multicolumn{1}{l|}{$\lambda$} & MAE            & RMSE            & MAPE            & $\rho$        & SNR             \\ \hline
1                          & 9.520          & 14.465          & 14.001          & 0.201          & -3.979          \\
0.8                        & 10.037         & 16.191          & 15.008          & 0.089          & -4.185          \\
0.5                        & \textbf{8.440} & \textbf{13.258} & \textbf{12.258} & \textbf{0.415} & \textbf{-2.736} \\
0.25                       & 8.641          & 14.558          & 13.137          & 0.256          & -3.056          \\ \hline
\end{tabular}%
}
\caption{Ablation study on the hyperparameter $\lambda$ in the loss functions used during the rPPG estimation stage. HR estimation results on the RPED dataset.}
\label{tab:ablation-lambda}
\end{table}

\subsection{Cross-dataset evaluation}
For cross-dataset evaluation, we follow the same protocol as RhythmFormer, where each dataset is split into 80 \% for training and 20 \% for validation, while the entire target dataset is used for testing.
This evaluation protocol effectively assesses the generalization capability of deep learning–based rPPG models under distribution shifts.
As shown in Table~\ref{tab:cross-data-1} and Table~\ref{tab:cross-data-2}, Periodic-MAE achieves substantial improvements across multiple evaluation metrics on the public benchmark datasets.
Compared with prior methods, our approach consistently demonstrates superior performance, indicating improved robustness and generalization in rPPG estimation.
Notably, although trained on the relatively simple PURE and UBFC-rPPG datasets, the proposed method achieves state-of-the-art performance on the more challenging MMPD and V4V benchmarks.
Across both cross-dataset settings, Periodic-MAE demonstrates substantial performance gains over RhythmFormer on the V4V benchmark, with MAE reduced by up to 42.5 \% and RMSE by up to 38.5 \%, highlighting its strong generalization capability.
The results suggest that self-supervised representation learning guided by physiological signal characteristics effectively enhances generalization to unseen data distributions.

To rigorously evaluate performance under challenging cross-dataset settings, we further conduct experiments on the RPED dataset, which is not used during the self-supervised pre-training stage.
The results are reported in Table~\ref{tab:cross-data-3}.
Notably, strong cross-dataset generalization is observed on the real-world dataset, collected in real-world environments under unconstrained lighting and subject motion.
Despite training solely on controlled laboratory data, Periodic-MAE achieves superior results compared to prior methods, demonstrating robust domain generalization.
The proposed method enables the model to learn data-agnostic rPPG representations that generalize well to real-world acquisition scenarios, even under substantially different data distributions.

In Figure~\ref{fig4}, the Bland-Altman plots for cross-dataset testing show results from a model trained on the PURE dataset and tested on the V4V test dataset.
The confidence interval of our model ranges from -17.946 to 20.133 bpm, while the confidence interval of RhythmFormer is from -25.496 to 33.335 bpm.
Additionally, Periodic-MAE demonstrates a stronger correlation with PPG HR values across the entire range compared to the previous method.
These findings indicate that our method responds effectively to a variety of subjects and environments.
Figure~\ref{fig4} also shows that the proposed model successfully estimates the rPPG signal, with peaks that closely align with those of the real PPG signal.
Since peak values are crucial for calculating HR, accurately matching the peaks between rPPG and PPG signals indicates the effectiveness of the model in capturing HR information.
We demonstrated that the proposed method effectively learns generic representations, enhancing the robustness of the rPPG estimation model.

\subsection{Intra-dataset evaluation}
We conducted intra-dataset testing on the PURE, UBFC-rPPG, MMPD, V4V, and RPED datasets, as shown in Table~\ref{intra-data} and Table~\ref{intra-data2}.
Our proposed method outperformed previous methods, indicating that the model effectively captures rPPG signals from facial videos.
In addition to public benchmarks, we evaluate intra-dataset performance on the real-world RPED dataset.
As shown in Table~\ref{intra-data2}, Periodic-MAE consistently outperforms previous methods across all metrics.
Compared with RhythmFormer, our approach reduces MAE by 25.0 \% and RMSE by 15.4 \% and achieves 71.1 \% higher SNR.
In particular, these gains are achieved using only unlabeled facial videos from other datasets during pre-training.
This demonstrates that physiology-guided self-supervised learning enables the model to acquire robust and transferable rPPG representations.
The learned features remain highly effective when fine-tuned for downstream tasks, even under real-world conditions that differ substantially from the pre-training distribution.

\subsection{Ablation study}
In this section, we present the results of an ablation study for a deeper analysis of the proposed method.
We evaluated the impact of masking strategies and frequency-domain constraints used during pre-training.
The experimental settings were the same as in the cross- and intra-dataset testing, and we fixed the random seed to ensure reproducibility.

\textbf{Impact of masking strategy}.
To demonstrate the effectiveness of the proposed masking strategy, we conducted an ablation study comparing it with previous masking methods introduced in VideoMAE.
Table~\ref{tab:4} presents the performance of models trained with different masking strategies and evaluated on downstream tasks. 
For the previous method, we set the masking ratio to 0.75 as used in~\citep{shao2023tranphys}.
Random frame masking provides excessive cues from adjacent frames, causing the model to overfit to the reconstruction task by exploiting simple frame-to-frame continuity rather than learning meaningful representations. 
Tube masking, while effective for capturing spatial features in general video understanding, overly emphasizes spatial continuity.
Since rPPG estimation requires learning pulse periodicity rather than spatial patterns, this spatial bias limits its effectiveness for physiological signal modeling.
In contrast, the proposed periodic masking strategy achieves an optimal balance between temporal context and task difficulty.
By dynamically adjusting the step size in each epoch, it prevents trivial solutions while encouraging the model to learn rhythmic representations that capture the periodic nature of physiological signals.
This design enables the model to focus on physiologically relevant facial regions and their temporal patterns during pre-training, leading to superior performance on downstream rPPG estimation tasks.

\textbf{Impact of loss function}.
The effectiveness of loss functions used in the pre-training stage is demonstrated in Table~\ref{tab:5}.
The reconstruction loss $L_r$ is used by default, as it is essential for the reconstruction task.
We evaluated the impact of additional losses, designed based on the assumptions of physiological bandlimits and sparsity of the power spectrum.
When using only the bandwidth constraint loss $L_b$, performance degraded as it merely restricted the bandwidth range.
Since cardiac activity generates a periodic signal, it exhibits distinct power components in the frequency domain.
We utilized $ L_s$ in the MAE framework to suppress noise signals with weak periodicity and avoid wideband prediction.
As shown in Table~\ref{tab:5}, optimizing the pre-training stage with frequency domain constraints effectively guides the model to focus on pulse signal representation in the pretext task.
Notably, we demonstrated that even without using additional spatio-temporal maps for reconstruction~\citep{shao2023tranphys, liu2024rppg}, our method provides effective cues about the periodic signal.

\textbf{Impact of hyperparameter \textbf{$\lambda$}}.
Table~\ref{tab:ablation-lambda} presents an ablation study on the weighting factor $\lambda$ in the rPPG estimation loss, evaluated on the RPED dataset. 
When $\lambda = 1$, performance degrades across all metrics, indicating that an imbalanced loss weighting disrupts the trade-off between temporal consistency and frequency alignment.
The best performance is achieved at $\lambda = 0.5$, yielding optimal results in terms of MAE, RMSE, MAPE, correlation coefficient, and SNR. 
This setting effectively leverages the pre-trained representations by jointly preserving temporal trends via $L_{time}$ and capturing cardiac periodicity while improving noise robustness through $L_{freq}$.
Overall, these results highlight the importance of properly balancing loss components during fine-tuning.

\section{Conclusion}
We proposed a method that learns data-agnostic physiological representations by leveraging a self-supervised learning approach to estimate rPPG.
The use of periodic masking and additional losses based on physiological bandlimits enhances the expressive encoding of pulse signals during pre-training.
Our method achieved promising results in cross-dataset testing, a particularly challenging task in heart rate estimation.
These findings indicate that the improved representation through pre-training contributed to the generalization of the rPPG estimation model.
Future work will involve testing on real-world datasets with more challenging environments and exploring additional downstream tasks, such as blood pressure and respiratory rate estimation.

\section*{Acknowledgements}
This work was supported by the Institute of Information \& Communications Technology Planning \& Evaluation(IITP)-Innovative Human Resource Development for Local Intellectualization program grant funded by the Korea government(MSIT) (IITP-2025-RS-2024-00439292) 

All data collection procedures were conducted under the approval of the Institutional Review Board (IRB Number: JBNU 2025-04-035-002). The authors would like to thank Juha Park for her assistance with data collection. 

\section*{Declaration of competing interest}
The authors declare that they have no known competing financial interests or personal relationships that could have appeared to influence the work reported in this paper.

\section*{CRediT authorship contribution statement}
\textbf{Jiho Choi:} Conceptualization, Methodology, Software, Validation, Data Curation, Writing – original draft, Writing - Review \& Editing.
\textbf{Sang Jun Lee:} Funding acquisition, Supervision, Writing – review \& editing.

\bibliographystyle{elsarticle-num}
\bibliography{references}

\end{document}